**Fairness Hub Technical Briefs**

# Towards a Consistent Measure of Model Bias

Jinsook Lee, Chris Brooks, Renzhe Yu, Rene Kizilcec

To measure bias, we encourage teams to consider using **AUC Gap**: the absolute difference between the highest and lowest test AUC for subgroups (e.g., gender, race, SES, prior knowledge). It is agnostic to the AI/ML algorithm used and it captures the disparity in model performance for any number of subgroups, which enables non-binary fairness assessments such as for intersectional identity groups.

## Measuring Bias using AUC Gap

The teams use a wide range of AI/ML models in pursuit of a common goal of doubling math achievement in low-income middle schools. Ensuring that the models, which are trained on datasets collected in many different contexts, do not introduce or amplify biases is important for achieving the goal. We offer here a versatile and easy-to-compute measure of model bias for all teams in order to create a common benchmark and an analytical basis for sharing what strategies have worked for different teams.

***AUC Gap*** [1] assumes that the goal is to make the most accurate predictions for all subgroups. Fairness is achieved when all subgroups have equal prediction performance. Unlike many other metrics [2, 3], AUC Gap works for any number of subgroups (e.g., intersectional gender-race groups, or multiple levels of SES and prior knowledge) and does not favor one prediction outcome over another (e.g., in other contexts, bias is associated with underpredicting a favorable outcome, such as release on bail or qualifying for a loan, for a disadvantaged group).

***Definition:*** AUC Gap is the largest difference between subpopulation AUCs (for the formal definition, see technical appendix below). It is a measure of the worst-case performance gap between a set of subpopulations. AUC Gap quantifies the largest disparity in predictive performance over subgroups and can be visualized along with AUC to add context to model results, as shown below in Figure 1.

***Uses and limitations:*** AUC Gap can be calculated for any prediction model, including ones that involve LLMs, so long as an AUC value can be derived: for multiclass prediction models, this can be achieved by predicting one vs. all other classes; for real number predictions, this can be achieved by thresholding. However, AUC Gap does not measure the source of bias that might exist (e.g. cannot measure bias in natural language, such as responses from a chatbot). We will provide guidance to teams on measuring bias in natural language in the coming weeks.

***Reporting:*** The Fairness Hub is funded until the end of 2024 to support teams. Our goal is to provide teams with resources for bias analysis and mitigation without the need for data sharing. We encourage all teams to compute AUC Gap for various models in their application and for various subgroups. We kindly ask that you share your results with our Hub so that we can assemble quantitative results from all teams about the biases they have identified.

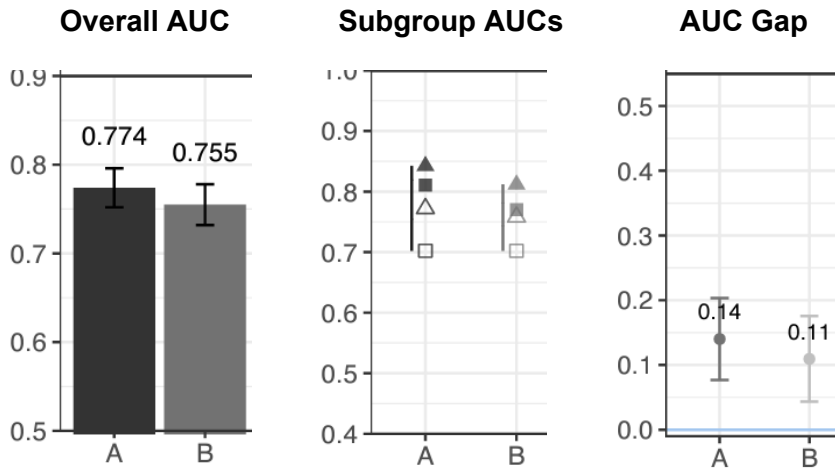

**Figure 1.** Example of visualizing the test AUC (left) alongside the subgroup test AUCs (center) and the AUC Gap (right) for four subgroups and two different models. While the subgroup AUCs (middle image) show the results for multiple groups at once (shown by triangles and squares which are filled in or not), it can be overwhelming when many combinations are compared. The AUC Gap (far right) provides a single statistic to capture this range. Source: [1].

# References

[1] Gardner, J., Yu, R., Nguyen, Q., Brooks, C., & Kizilcec, R. (2023). Cross-Institutional Transfer Learning for Educational Models: Implications for Model Performance, Fairness, and Equity. In *Proceedings of the 2023 ACM Conference on Fairness, Accountability, and Transparency* (pp. 1664-1684).

[2] Kizilcec, R. F., & Lee, H. (2022). Algorithmic fairness in education. In The ethics of artificial intelligence in education (pp. 174-202). Routledge.

[3] Hort, M., Chen, Z., Zhang, J. M., Sarro, F., & Harman, M. (2022). Bias mitigation for machine learning classifiers: A comprehensive survey. *arXiv preprint arXiv:2207.07068*.

# Technical Appendix

The Area Under the Receiver Operating Characteristic Curve (AUC) is defined as:

$$AUC(f(\theta)) = \int_0^1 TPR(FPR(f_t(\theta)))dt \qquad (1)$$

where $t$ indicates a prediction threshold applied to the predictions of the model (i.e. using the decision rule $f_t(\theta, x) = \mathbb{1}{f(\theta, x) \geq t}$), and TPR, FPR are the true positive rates and false positive rates. AUC scores are constrained to $[0, 1]$, with a random predictor achieving an AUC of 0.5.

We define the **AUC gap** as the largest disparity in predictive performance across a set of arbitrarily many (possibly-overlapping) subgroups $G$. Formally, the AUC Gap is calculated as:

$$max_{g,g' \in G} |E_{D_k}[f(\theta(D_{k,g}))] - E_{D_k}[f(\theta(D_{k,g'}))]| \qquad (2)$$

where $D_{k,g}$ and $D_{k,g'}$ indicate the subset of the data in group $g$ and $g'$, respectively.